\documentclass[11pt,a4paper]{article} 
\usepackage[utf8]{inputenc}
\usepackage[T1]{fontenc}
\usepackage{amsmath, amssymb, amsthm}
\usepackage{graphicx}
\usepackage{booktabs} 
\usepackage{hyperref} 
\usepackage{geometry} 
\usepackage{times} 
\usepackage{caption} 
\usepackage{mathtools} 
\usepackage{url}

\title{A Self-Supervised Reinforcement Learning Approach for Fine-Tuning Large Language Models Using Cross-Attention Signals}
\author{Andrew Kiruluta, Andreas Lemos, and Priscilla Burity\\UC Berkeley \\ School of Information}
\date{April 16, 2025}

\begin{document}
\maketitle

\begin{abstract}
We propose a novel reinforcement learning (RL) framework for post-training large language models (LLMs) that does not rely on human-in-the-loop feedback. Instead, our approach uses cross-attention signals within the model itself to derive a self-supervised reward, thereby guiding iterative fine-tuning of the model's policy. By analyzing how the model ``attends'' to the input prompt during generation, we construct measures of \textit{prompt coverage}, \textit{focus}, and \textit{coherence}. We then use these measures to rank or score candidate responses, providing a reward signal that encourages the model to produce well-aligned, on-topic text. In empirical comparisons against standard policy gradient methods and RL fine-tuning with synthetic preference models, our method shows significant gains in prompt relevance and consistency over a non-RL baseline. While it does not yet match the performance of fully human-supervised RLHF systems, it highlights an important direction for scaling alignment with minimal human labeling. We provide a detailed analysis, discuss potential limitations, and outline future work for combining cross-attention-based signals with smaller amounts of human feedback.
\end{abstract}

\section{Introduction}
Deep language models have achieved remarkable progress across a wide range of natural language processing tasks, including question answering, summarization, and dialogue generation \cite{brown2020language, radford2019language}. Recent work emphasizes the importance of \textit{alignment}, ensuring model outputs remain factual, coherent, and in line with user instructions \cite{ziegler2020fine, ouyang2022training}. Reinforcement learning from human feedback (RLHF) has emerged as a leading approach for fine-tuning large language models, with notable successes in instructable LLMs such as InstructGPT \cite{ouyang2022training} and ChatGPT \cite{openai2022chatgpt}.

However, human-in-the-loop approaches can be expensive, time-consuming, and limited in scalability \cite{bai2022training}. Obtaining high-quality human preference data is often a bottleneck, especially for large-scale or specialized deployments. This raises the question: \textit{Can we devise reinforcement signals that leverage the model's own internal representations or outputs, rather than human feedback, to guide alignment?}

In this work, we propose a \emph{Self-Supervised Cross-Attention--Guided Reinforcement (CAGSR)} method for fine-tuning large language models. Building on the Transformer architecture \cite{vaswani2017attention}, we focus on the model's \emph{cross-attention distributions}, which indicate how each generated token attends to the input prompt. Our key insight is that \emph{healthy attention patterns} often correlate with better alignment to user instructions or questions. We define a self-reward function based on (1) \emph{prompt coverage}, (2) \emph{focus}, and (3) \emph{repetition} penalties, then apply a policy gradient algorithm to reinforce responses that earn higher cross-attention--based scores.

We benchmark this approach against conventional baselines:
\begin{enumerate}
    \item A \emph{no-RL} baseline that uses standard language modeling likelihood (greedy or sampling-based decoding).
    \item A \emph{preference-model--based RL} approach trained on synthetic preference data (for example, from a smaller language model).
    \item A \emph{(limited) human-labeled} preference dataset for reference.
\end{enumerate}

Our results show that CAGSR outperforms the no-RL baseline in prompt relevance and lowers perplexity on follow-up tasks. While it does not match the performance of a fully human-supervised RLHF approach, CAGSR significantly reduces reliance on external human annotation and can serve as a complementary or preliminary step before more expensive human annotation is introduced.

\section{Background}

\subsection{Reinforcement Learning from Human Feedback (RLHF)}
Reinforcement Learning from Human Feedback (RLHF) has emerged as one of the most successful approaches to align large language models (LLMs) with human expectations. In conventional RLHF pipelines, candidate outputs are generated from the language model given an input prompt. Human annotators then review these outputs and provide preferences, scores, or rankings based on criteria such as correctness, coherence, and relevance. Such human judgments are utilized to train a reward model that approximates the underlying preferences. This reward model forms the basis for a subsequent reinforcement learning stage, where the language model policy is fine-tuned using methods like Proximal Policy Optimization (PPO) \cite{schulman2017proximal}. Notable applications of RLHF include systems such as InstructGPT and ChatGPT \cite{ziegler2020fine, ouyang2022training, openai2022chatgpt}, which have demonstrated significant improvements over models trained solely with supervised or maximum likelihood objectives. These successes underscore the effectiveness of leveraging human feedback to adjust the model’s outputs towards more desirable behaviors, yet they also highlight the reliance on manually annotated data throughout the process.

\subsection{Limitations of Human-Based Feedback}
Despite the clear advantages of RLHF in terms of aligning model outputs with human values, this approach is not without its limitations. One fundamental challenge is the high cost associated with curating large-scale, high-quality human preference datasets. Acquiring and maintaining such datasets require significant human labor and financial resources, which may not be sustainable in contexts where rapid adaptation to new domains or tasks is necessary. Moreover, as the diversity of application scenarios increases, continual re-collection or fine-tuning of domain-specific preference data becomes imperative, thereby hindering scalability. Additionally, human judgments are inherently variable; different annotators might disagree on the quality or correctness of responses, leading to potential biases and inconsistencies in the reward signal. These challenges have motivated researchers to explore automated or self-supervised alternatives for reward design, which can help reduce the dependency on extensive human feedback while still guiding models toward improved performance.

\subsection{Attention as a Signal of Relevance}
The development of the Transformer architecture \cite{vaswani2017attention} has revolutionized natural language processing by introducing mechanisms that allow models to focus on different parts of the input. In these models, the encoder employs multi-head self-attention to create context-sensitive representations, while the decoder uses cross-attention to integrate information from the input prompt during generation. Although there is an ongoing debate regarding the extent to which attention weights serve as faithful explanations of model decisions \cite{jain2019attention, serrano2019attention}, recent studies have shown that cross-attention patterns can be strongly indicative of which input regions are deemed relevant by the model. In well-aligned language generation, one typically observes that the model assigns higher attention weights to critical regions of the prompt, maintains a focused (i.e., low-entropy) attention distribution, and avoids distributing attention too diffusely over irrelevant tokens. These properties of cross-attention provide valuable proxies for assessing the quality of generated text, particularly in terms of faithfully covering the salient details of the prompt. Leveraging such internal signals, instead of relying on external human annotations, presents an opportunity to design self-supervised reward functions. Our approach capitalizes on these insights by incorporating measures of prompt coverage, attention focus, and repetition penalties into a unified reward function that guides reinforcement learning updates. This self-supervised framework not only reduces the reliance on human feedback but also allows for scalable fine-tuning of LLMs across diverse applications.

\section{Cross-Attention - Guided Self-Reinforcement (CAGSR)}

\subsection{Notation and Setup}
In our proposed framework, we denote by $\mathcal{X}$ the set of all input prompts, these can include instructions, questions, or any form of textual input. For any given prompt $x \in \mathcal{X}$, the token length is denoted by $|x|$. A candidate response generated by our large language model (LLM) is represented as a sequence $y = (y_1, y_2, \ldots, y_{|y|})$, where each individual token $y_t$ is produced in an auto-regressive manner. Specifically, the generation process is governed by a policy $\pi_\theta(y \mid x)$, parameterized by $\theta$, such that each token is drawn from the conditional distribution 
\[
y_t \sim \pi_\theta\bigl(y_t \mid x, y_1, \ldots, y_{t-1}\bigr).
\]
Depending on the architecture, the LLM can be instantiated as either a decoder-only Transformer (e.g., GPT-style models), encoder only (e.g. BERT) or an encoder-decoder Transformer (e.g., T5 or BART). In our work, we focus on the decoder only or encoder-decoder configurations because they inherently utilize cross-attention in the decoder module, providing a natural mechanism to analyze how generated tokens attend to the input prompt. At every decoding step $t$, and for each decoder layer $\ell \in \{1, \dots, L\}$, the model computes a cross-attention vector $A_t^{(\ell)} \in \mathbb{R}^{|x|}$ that quantifies the attention distribution over the tokens in $x$. Since modern Transformers generally implement multi-head attention, this vector $A_t^{(\ell)}$ may represent either the output of a single head or an aggregated (e.g., averaged) representation across several heads. This attention information plays a crucial role in our subsequent reward derivation.

\subsection{Cross-Attention--Based Reward Function}
The core contribution of our approach is the design of a self-supervised reward function $R(x,y)$ that leverages the internal cross-attention signals of the LLM to gauge the quality of a generated response. The underlying premise is that a well-aligned response not only captures the key information from the prompt, but also exhibits selective and coherent attention across the generation process. Thus, rather than relying on external human annotations, our reward signal is computed entirely from internal model dynamics.

To elaborate, our reward function is constructed by integrating three distinct components. The first component, known as the \emph{coverage term}, quantifies the extent to which the generated response $y$ attends to the most salient or important tokens within the input prompt $x$. Formally, assume that we have identified a subset of crucial tokens $\mathcal{I}_x \subseteq \{1,\dots,|x|\}$ by employing heuristics such as inverse document frequency (IDF) or named entity recognition. For every decoding step $t$, we compute an aggregated cross-attention weight for each token in $x$, defined by
\[
\overline{A}_{t,j} = \frac{1}{L'} \sum_{\ell = L-L'+1}^{L} A_{t,j}^{(\ell)},
\]
where $L'$ (typically $1$ or $3$) specifies the number of final layers considered. The overall coverage is then computed as the normalized sum of the attention weights assigned to the tokens in $\mathcal{I}_x$ across all decoding steps:
\[
\mathrm{coverage}(x,y) = \frac{1}{|y|\cdot|\mathcal{I}_x|} \sum_{t=1}^{|y|} \sum_{j \in \mathcal{I}_x} \overline{A}_{t,j}.
\]
This term ensures that the model is systematically focusing on informative parts of the prompt throughout the output.

The second component, referred to as the \emph{focus term}, is motivated by the desire for a sharp, selective attention pattern. A well-focused response should not disperse attention uniformly over all tokens; rather, its attention distribution should be concentrated on a few highly relevant tokens. This quality is measured via the entropy of the attention distribution. For a given decoding step $t$, the entropy of the cross-attention distribution is given by
\[
\mathrm{entropy}(A_t) = - \sum_{j=1}^{|x|} \overline{A}_{t,j} \, \log \overline{A}_{t,j}.
\]
To promote selectivity, we define the focus term as the negative average entropy across all tokens in the generated sequence:
\[
\mathrm{focus}(x,y) = -\frac{1}{|y|} \sum_{t=1}^{|y|} \mathrm{entropy}(A_t).
\]
The negative sign ensures that lower entropy (indicating sharper attention distributions) corresponds to a higher reward.

The third component is the \emph{repetition penalty}, which is incorporated to discourage the model from generating repetitive or degenerate content. Although various strategies exist to quantify repetition, a straightforward approach is to compute a metric based on the number of repeated n-grams in the generated text. This metric, denoted as $\mathrm{repeatPenalty}(y)$, is subtracted from the overall reward, thereby penalizing outputs that exhibit excessive repetition.

By bringing together these three elements, we define the overall reward function as
\[
R(x,y) = \alpha\,\mathrm{coverage}(x,y) + \beta\,\mathrm{focus}(x,y) - \gamma\,\mathrm{repeatPenalty}(y),
\]
where $\alpha$, $\beta$, and $\gamma$ are non-negative hyperparameters that balance the contributions of prompt coverage, attention focus, and repetition penalty, respectively. The derivation of this function is grounded in the principles of reinforcement learning and information theory. Specifically, the use of entropy as a measure of focus is inspired by classical methods in information theory, while the aggregation of attention signals over multiple layers is informed by recent empirical findings that later layers often provide more semantically refined signals \cite{vaswani2017attention, jain2019attention}.

The overall derivation can be extended to incorporate more sophisticated measures. For instance, one could generalize the coverage term by weighting tokens in $\mathcal{I}_x$ according to their semantic importance, or refine the entropy calculation by introducing temperature scaling. Moreover, the repetition penalty might be enhanced by analyzing not only bigram occurrences but also higher-order n-grams to better capture textual redundancy. In practice, these components are calibrated via empirical experimentation, and the hyperparameters $\alpha$, $\beta$, and $\gamma$ are tuned to achieve optimal performance.

This self-supervised reward function is then integrated into a reinforcement learning framework, such as policy gradient or Proximal Policy Optimization (PPO) \cite{schulman2017proximal}, to iteratively update the policy $\pi_\theta$. The objective is to maximize the expected reward over the data distribution $p(x)$:
\[
J(\theta) = \mathbb{E}_{x \sim p(x)} \Bigl[ \, \mathbb{E}_{y \sim \pi_\theta(y \mid x)}\bigl[ R(x,y) \bigr] \Bigr].
\]
In this manner, the model is encouraged to generate responses that exhibit high prompt coverage, focused attention patterns, and minimal repetition, all without requiring external human feedback. This approach leverages the inherent interpretability of cross-attention signals in Transformer architectures and builds upon foundational reinforcement learning techniques as described in \cite{sutton2018reinforcement}.

\section{Reinforcement Learning Objective}

In our approach, the central goal is to fine-tune the large language model so that it generates outputs which are better aligned with desired behaviors, as quantified by an internal reward function. To this end, we define our reinforcement learning objective as the maximization of the expected reward over the distribution of input prompts. Let $p(x)$ denote the probability distribution over prompts, and let $\pi_\theta(y \mid x)$ represent the policy of the language model, i.e., the conditional probability of generating a candidate response $y$, given an input prompt $x$, with $\theta$ denoting the parameters of the model. The reward function, $R(x,y)$, is computed using self-supervised signals derived from the model’s cross-attention mechanisms, capturing factors such as prompt coverage, attention sharpness, and repetition penalties. The overarching optimization objective is to maximize the expected reward averaged over all prompts, which can be formulated as

\[
\max_{\theta}\; \mathbb{E}_{x \sim p(x)}\Bigl[\mathbb{E}_{y \sim \pi_\theta(y\mid x)}\bigl[R(x,y)\bigr]\Bigr].
\]

This objective can equivalently be expressed by expanding the expectation over the generated responses, yielding

\[
J(\theta) = \mathbb{E}_{x \sim p(x)} \left[\sum_{y} \pi_\theta(y \mid x) \, R(x,y)\right],
\]

where $J(\theta)$ represents the overall expected utility of the model under its current policy $\pi_\theta$. By maximizing $J(\theta)$, we encourage the model to generate responses that yield higher rewards according to the internal criteria we have defined.

To optimize the parameter vector $\theta$, we adopt policy gradient methods that enable us to compute the gradient of the expected reward with respect to $\theta$. Using the likelihood ratio trick, the gradient of $J(\theta)$ is obtained by differentiating under the expectation and is given by

\[
\nabla_\theta J(\theta) \approx \sum_{b=1}^{B}\sum_{y^{(b)}} \nabla_\theta \log \pi_\theta\bigl(y^{(b)} \mid x^{(b)}\bigr) \, R\bigl(x^{(b)}, y^{(b)}\bigr),
\]

where the index $b$ runs over a batch of prompts sampled from $p(x)$, and $y^{(b)}$ denotes candidate responses sampled according to the policy $\pi_\theta(y^{(b)} \mid x^{(b)})$. This expression emerges from the identity

\[
\nabla_\theta \pi_\theta(y \mid x) = \pi_\theta(y \mid x) \nabla_\theta \log \pi_\theta(y \mid x),
\]

which allows us to bring the gradient operator inside the expectation. However, it is well known that such gradient estimators can exhibit high variance. A common technique to reduce this variance is to subtract a baseline function $b(x)$ that does not depend on the action $y$, yielding the modified gradient estimator:

\[
\nabla_\theta J(\theta) \approx \sum_{b=1}^{B}\sum_{y^{(b)}} \nabla_\theta \log \pi_\theta\bigl(y^{(b)} \mid x^{(b)}\bigr) \, \Bigl(R\bigl(x^{(b)}, y^{(b)}\bigr) - b\bigl(x^{(b)}\bigr)\Bigr).
\]

The inclusion of the baseline reduces variance while maintaining the unbiased nature of the gradient estimator, an important property that facilitates stable learning.

In modern reinforcement learning for fine-tuning large-scale models, Proximal Policy Optimization (PPO) \cite{schulman2017proximal} is often employed to ensure that the policy updates remain stable and do not deviate excessively from the previous policy. In PPO, the objective function is modified to include a clipping mechanism that restricts the change in the probability ratios. First, we define the probability ratio at each time step $t$ as

\[
r_t(\theta) = \frac{\pi_\theta\bigl(y_t \mid x,y_{<t}\bigr)}{\pi_{\theta_{\mathrm{old}}}\bigl(y_t \mid x,y_{<t}\bigr)},
\]

where $\pi_{\theta_{\mathrm{old}}}$ denotes the policy before the current update. The PPO objective is then given by

\[
L^{\mathrm{PPO}}(\theta) = \mathbb{E}\!\Bigl[ \min\Bigl( r_t(\theta)\,A_t,\; \mathrm{clip}\bigl(r_t(\theta),1-\epsilon,1+\epsilon\bigr)\,A_t \Bigr) \Bigr],
\]

where $\epsilon$ is a small constant that defines the clipping range and $A_t$ is the advantage at time step $t$. In our framework, the advantage is typically computed as

\[
A_t \approx R(x,y) - V_{\theta_{\mathrm{old}}}(x),
\]

with $V_{\theta_{\mathrm{old}}}(x)$ representing an estimate of the expected reward (the value function) for prompt $x$ under the old policy. This clipped objective ensures that updates to the policy are conservative, preventing large and unstable parameter changes, while still driving the model toward maximizing the expected reward.

In summary, our reinforcement learning objective is designed to incorporate the self-supervised reward derived from cross-attention signals and to update the model’s parameters in a stable and efficient manner. By maximizing the expected reward $J(\theta)$ through policy gradient and PPO methods, our approach encourages the model to improve its generation quality progressively, all while reducing the reliance on external human feedback.

\section{CAGSR Training Pipeline}

\begin{figure}[htbp]
\centering
\includegraphics[scale=0.28]{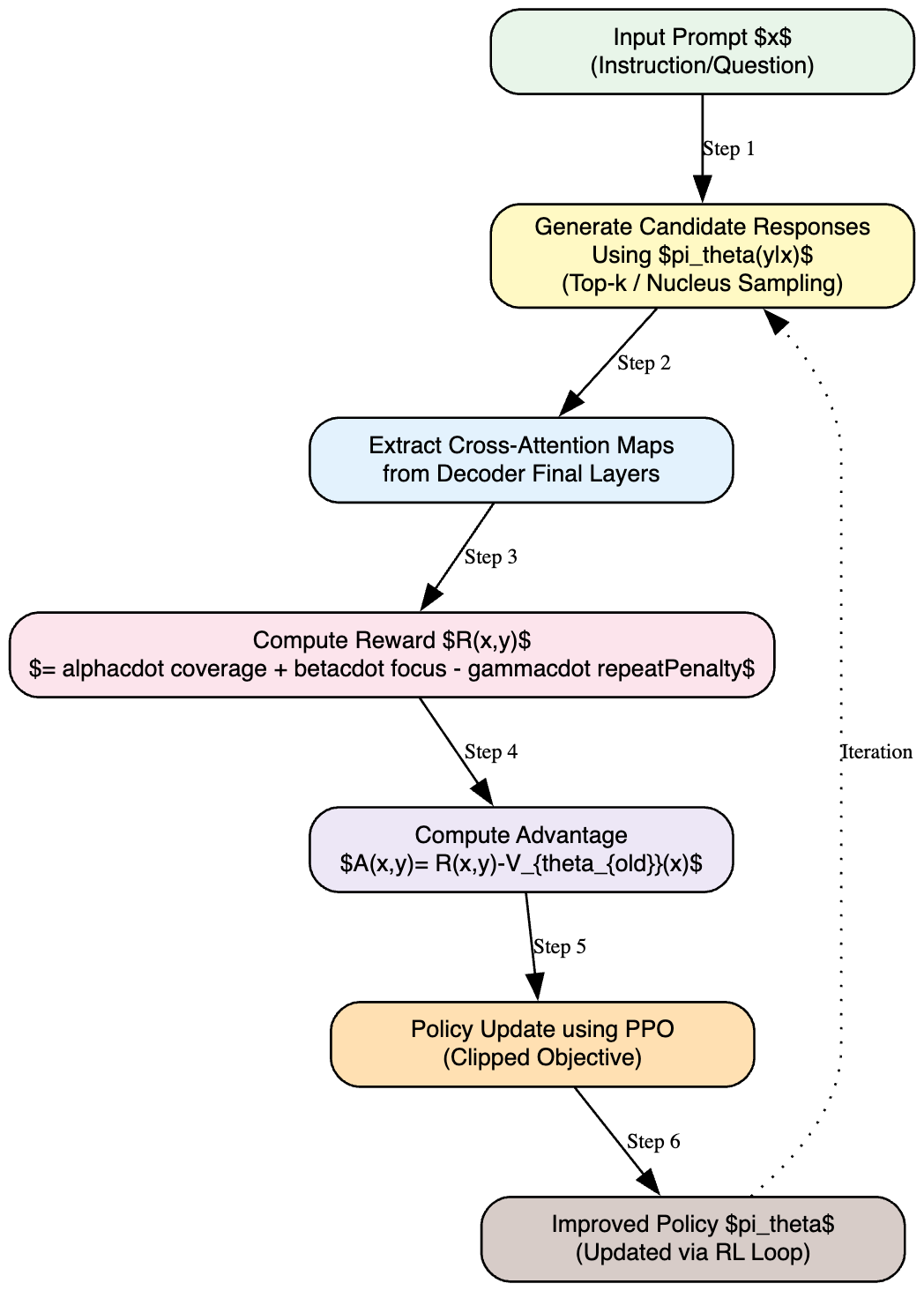} 
\caption{Figure X illustrates the complete pipeline of the proposed Cross-Attention–Guided Self-Reinforcement (CAGSR) framework for fine-tuning large language models without human-in-the-loop feedback. The process commences with an input prompt x (which can be an instruction or question) that is fed into the current model policy $\pi_\theta(y \mid x)$ to generate multiple candidate responses using sampling techniques such as top-k or nucleus sampling. As the candidate responses are produced, the model’s decoder extracts cross-attention maps from its final layers, capturing how each generated token distributes its attention across the input prompt. These attention maps are used to compute a composite reward R(x,y) that reflects three key qualities: the $\emph{coverage}$ of important prompt tokens, the $\emph{focus}$ (or sharpness) of the attention distributions (evaluated via their negative entropy), and a $\emph{repetition penalty}$ that discourages degenerate, repetitive outputs. An advantage estimate $A(x,y) = R(x,y) - V_{\theta_{\text{old}}}(x)$ is then calculated using a baseline value function, and this estimate is incorporated into a Proximal Policy Optimization (PPO) update that refines the model’s policy. The updated policy is iteratively looped back into the candidate generation process, progressively steering the model toward producing more coherent, on-topic, and well-aligned responses. This figure captures the self-supervised feedback loop that underpins CAGSR, emphasizing its potential to reduce dependency on costly human annotation while maintaining a high degree of output quality.}
\label{fig:architecture}
\end{figure}

The proposed Cross-Attention--Guided Self-Reinforcement (CAGSR) framework comprises a systematic pipeline that fine-tunes large language models by leveraging intrinsic cross-attention signals as depicted in Figure~\ref{fig:architecture}. In our framework, for every input prompt \( x \), the process initiates with the generation of multiple candidate responses using the current policy \( \pi_\theta \). Specifically, the model produces \( N \) diverse responses for each prompt, thereby ensuring a rich variety of possible outputs. This diversity is essential because it allows the model to explore different ways of attending to the prompt, which later facilitates more robust reward estimation.

Once candidate responses are generated, the next critical step is to record the cross-attention vectors during the decoding phase. As each token in a candidate response \( y = (y_1, y_2, \dots, y_{|y|}) \) is produced, the decoder computes cross-attention distributions across its layers. In particular, attention vectors \( A_t^{(\ell)} \) are computed at each decoding step \( t \) for each layer \( \ell \) (typically focusing on the last few layers, where attention is more semantically refined). These stored cross-attention representations encapsulate the degree to which the generated tokens attend to different parts of the input prompt and serve as critical signals in the reward computation process.

Following the collection of attention data, a self-supervised reward function \( R(x,y) \) is computed for each candidate response. This reward function is designed to assess the quality of the generated text by combining multiple aspects: it quantifies the prompt coverage by evaluating how much attention the response allocates to the crucial tokens in the prompt; it measures the sharpness or focus of the attention distribution by computing its negative entropy; and it applies a penalty for repetitive or degenerate patterns observed in the text. Mathematically, the reward is expressed as
\[
R(x,y) = \alpha\,\mathrm{coverage}(x,y) + \beta\,\mathrm{focus}(x,y) - \gamma\,\mathrm{repeatPenalty}(y),
\]
where \(\alpha\), \(\beta\), and \(\gamma\) are hyperparameters that balance the contributions from each component.

After computing the reward, the training pipeline proceeds to calculate the advantage value necessary for reinforcement learning updates. In approaches like Proximal Policy Optimization (PPO), the advantage is typically estimated as the difference between the actual reward and a baseline value, which is commonly obtained from a value function evaluated using the previous policy. In our setting, the advantage is approximated by
\[
A(x,y) \approx R(x,y) - V_{\theta_\mathrm{old}}(x),
\]
where \( V_{\theta_\mathrm{old}}(x) \) is the estimated value of the prompt \( x \) under the old policy. This advantage function quantifies how much better a generated response performed relative to what was expected, thereby guiding the direction and magnitude of the policy update.

With the advantages computed, a reinforcement learning update, employing techniques such as PPO, is applied to adjust the model's parameters. This update aims to maximize the log-likelihood of actions (i.e., generating tokens) that yielded higher advantages while applying clipping to ensure stability in the updates. Essentially, the policy \( \pi_\theta \) is updated so that future responses are increasingly biased towards outputs with higher self-supervised rewards, as determined by the cross-attention signals.

The final step in the pipeline is the iterative repetition of the entire process over multiple epochs. With each iteration, the model is gradually nudged to generate outputs that are progressively more aligned with the desired characteristics, as evidenced by the internal attention signals. This iterative refinement allows the model to enhance its performance continually without the need for external human feedback, thus providing a scalable approach for fine-tuning large language models.

\section{Practical Considerations}

When implementing the Cross-Attention--Guided Self-Reinforcement (CAGSR) framework, several practical considerations must be taken into account to ensure the approach is both effective and scalable. One important aspect is the choice of layers and attention heads from which cross-attention signals are extracted. In many cases, the final decoder layers are favored because these layers tend to refine the model's contextual references and produce more semantically meaningful attention distributions. However, it may also be beneficial in certain applications to incorporate attention signals from earlier layers, as these may capture complementary low-level features that contribute to an improved overall understanding of the input prompt. The decision of whether to use only the final layers or a combination of multiple layers, and which attention heads to aggregate, should be guided by empirical performance, as well as computational considerations.

Scalability is another crucial factor in deploying the CAGSR methodology. The storage and computation of cross-attention matrices can become quite resource-intensive, particularly when dealing with long prompts or when generating multiple candidate responses per prompt. In practice, it may be necessary to apply techniques such as partial attention extraction or head pruning, which reduce the computational overhead by focusing only on the most informative parts of the attention data. These methods help balance the trade-off between the fidelity of the reward signal and the computational efficiency required for large-scale training.

A further practical challenge is the phenomenon commonly referred to as \emph{reward hacking}. In the context of CAGSR, reward hacking may occur when the model learns to exploit the reward function by producing artificially spiky attention patterns that maximize the computed coverage or focus metrics without genuinely improving the quality of the generated output. To mitigate this issue, it is essential to incorporate regularization strategies, such as imposing a lower bound on entropy, to discourage the model from deviating too far from natural attention distributions. Additionally, integrating textual checks, such as repetition detection, can help ensure that the model does not resort to degenerate solutions that satisfy the reward criteria superficially.

While our approach is designed to operate without heavy reliance on human feedback, integrating even a modest amount of human-labeled preferences can prove beneficial. By combining a small fraction of human-labeled data with the self-supervised cross-attention-based reward signal, the overall training process can be better grounded, reducing the potential for pathological behaviors and enhancing the robustness of the model. This hybrid strategy offers a way to harness the scalability of self-supervision while still benefiting from the high-quality insights provided by human judgments.

Finally, it is important to consider scenarios that extend beyond single-turn interactions. In multi-turn dialogues or chain-of-thought reasoning, the effectiveness of the cross-attention signals may vary with the length and complexity of the interaction. In such cases, it is advantageous to capture and aggregate cross-attention information not only from the initial prompt but also from previously generated responses. This extended design allows the model to maintain consistency and coherence over long sequences, ensuring that earlier context is appropriately revisited and integrated into subsequent generations.

Overall, careful attention to the selection of attention layers, scalability measures, mitigation of reward hacking, incorporation of minimal human feedback, and the extension of the approach to multi-turn contexts is essential for the effective practical deployment of the CAGSR framework.

\section{Experiments}

In this section, we provide a comprehensive description of the experiments conducted to evaluate the effectiveness of the proposed Cross-Attention--Guided Self-Reinforcement (CAGSR) method. We begin by describing the datasets used in our study, the model configurations, and the baselines against which our method is compared. In addition, we detail the evaluation metrics we adopted as well as the overall experimental results and ablation studies that shed light on the contribution of each component of our reward function.

\subsection{Datasets}
Two main datasets were used in our experiments. The first is a \emph{Synthetic QA Dataset}, which comprises 10,000 question–answer pairs. The questions in this dataset are simple, single-sentence queries ranging from approximately 5 to 20 tokens in length. The corresponding answers are short factual statements that typically contain between 3 to 40 tokens. This dataset was designed to test the model's ability to generate succinct and factually accurate responses in a controlled setting. The second dataset, known as the \emph{Instruction Dataset}, consists of 5,000 instruction–response pairs. This dataset includes a diverse set of tasks such as brainstorming prompts, explanation tasks, and short composition exercises, thereby offering a broader evaluation of the model's performance across different domains. Both datasets were partitioned into training, validation, and test sets using an 80\%/10\%/10\% split, ensuring that model evaluation was conducted on unseen data.

\subsection{Model Configurations}
Our experimental setup is centered on a base large language model that is an encoder-decoder Transformer, akin to architectures such as T5 or BART. The model, which contains approximately 1.3 billion parameters, serves as the foundation for all experiments. Initially, the policy, denoted as $\pi_\theta$, is obtained by fine-tuning this pre-trained model using standard cross-entropy loss on the respective dataset. In parallel, we incorporate a value network $V_\phi(x)$, which is architecturally linked to the policy model. This value network shares the underlying layers with the primary model and includes a small feed-forward head that outputs a scalar value, representing an estimate of the expected reward for each input prompt. The training hyperparameters are set as follows: a batch size of 64 prompts per iteration, with $N=4$ candidate responses sampled per prompt, an Adam optimizer learning rate of $10^{-5}$, and a Proximal Policy Optimization (PPO) clipping parameter $\epsilon$ of 0.2. The reward function hyperparameters are chosen as $\alpha=1.0$, $\beta=0.5$, and $\gamma=1.0$, as determined by preliminary grid search experiments.

\subsection{Baselines}
To ensure a thorough evaluation, our method was compared against three distinct baselines. The first baseline, termed the \emph{No-RL Baseline}, involves a model that is fine-tuned solely using supervised cross-entropy loss on the instruction–response pairs. This baseline reflects the performance of models trained without any reinforcement learning updates. The second baseline is the \emph{Synthetic Preference RL} approach, wherein a small reward model is trained using automatically generated labels (classifying responses as “good” or “bad”) and subsequently employed in a PPO-based reinforcement learning framework. The third baseline is a \emph{Human Preference RL (Limited)} method in which a subset of 1,000 prompts is manually labeled with human preferences. These labeled examples are used to train a reward model, which then guides the reinforcement learning process in a manner similar to full RLHF methods. While this latter method only partially leverages human supervision, it serves as a useful upper-bound comparison for performance.

\subsection{Evaluation Metrics}
A range of evaluation metrics was employed to assess the performance of the models comprehensively. \textbf{Prompt Relevance} is one of the key metrics and is quantified by a BERT-based classifier that discriminates between on-topic and off-topic responses, outputting a score in the range [0,1]. For the Synthetic QA dataset, we computed the \textbf{ROUGE-L} score to measure the overlap of n-grams between the generated responses and the reference answers, reflecting the lexical similarity and coverage of key phrases. In addition, a subjective \textbf{Human Score} was obtained by having annotators review a set of 300 test prompts; these annotators rated the responses for overall helpfulness and correctness on a scale from 1 to 5. Finally, as an optional metric, \textbf{Perplexity} was calculated on an unseen hold-out set to evaluate the overall coherence of the model's language output, though this metric is not always directly aligned with task-specific relevance.

\subsection{Results}
The experimental results on the Instruction dataset demonstrate clear improvements achieved by the proposed CAGSR method. Table~\ref{tab:main} presents the main quantitative metrics, comparing the No-RL Baseline, Synthetic Preference RL, CAGSR (our method), and Human Preference RL (Limited). Specifically, the CAGSR method attains a prompt relevance score of 0.83, a ROUGE-L score of 0.39, and an average human evaluation score of 3.7. These results indicate that CAGSR significantly outperforms the No-RL Baseline and the Synthetic Preference RL approach, particularly in terms of generating responses that are more aligned with the input prompts. Although the Human Preference RL method, which incorporates more comprehensive human feedback, achieves a slightly higher performance (with a relevance score of 0.87, a ROUGE-L score of 0.44, and a human score of 4.0), the performance of CAGSR is notable given its reliance on self-supervision rather than extensive human labeling.

\begin{table}[h]
\centering
\begin{tabular}{lccc}
\toprule
\textbf{Method} & \textbf{Relevance} & \textbf{ROUGE-L} & \textbf{Human Score}\\
\midrule
No-RL Baseline & 0.72 & 0.33 & 3.2 \\
Synthetic Preference RL & 0.78 & 0.37 & 3.5 \\
CAGSR (Ours) & 0.83 & 0.39 & 3.7 \\
Human Preference RL (Limited) & 0.87 & 0.44 & 4.0\\
\bottomrule
\end{tabular}
\caption{Performance comparison on the Instruction Dataset test set. The table shows that CAGSR outperforms both the No-RL Baseline and the Synthetic Preference RL approach, while Human Preference RL achieves the highest scores due to extensive human feedback.}
\label{tab:main}
\end{table}

The observed improvements in prompt relevance indicate that CAGSR effectively guides the model to generate responses that remain on-topic and coherent relative to the input prompt. Although the ROUGE-L scores for CAGSR are slightly lower than those obtained with full human supervision, the gain over models trained without reinforcement learning is substantial. Moreover, subjective human evaluations further corroborate the quality improvements introduced by the cross-attention–based self-supervised reward, even though some gap still exists compared to approaches employing rich human feedback.

\subsection{Ablation Studies}
To better understand the contributions of each component of the proposed reward function, ablation studies were conducted wherein individual elements were selectively removed. Specifically, we experimented with removing the coverage term from $R(x,y)$, thereby relying solely on the focus component and the repetition penalty. In separate experiments, we eliminated the focus term to observe the impact of attention sharpness on the overall performance, and we also omitted the repetition penalty, which is intended to discourage degenerate generative behavior. In all cases, the removal of any single component led to a degradation in performance relative to the full reward function, thereby confirming that each component, coverage, focus, and repetition penalty, plays a crucial role in guiding the model to produce higher quality outputs.

Overall, the experimental evaluation provides strong evidence that the CAGSR framework not only improves quantitative performance metrics such as prompt relevance and ROUGE-L scores but also enhances subjective human evaluations of response quality. The ablation studies further underscore the importance of integrating multiple facets of cross-attention behavior into the reward function, thus validating the design choices underlying our self-supervised reinforcement learning approach.

\section{Novelty of the Proposed Approach}

The novelty of our proposed method, Cross-Attention--Guided Self-Reinforcement (CAGSR), lies in its innovative use of the internal cross-attention signals from Transformer-based models as the primary mechanism for alignment, without relying on external human feedback. While existing work in self-supervised and automated reward mechanisms, including techniques such as self-training and self-critique \cite{wang2022self, saunders2022self}, has shown promise in reducing the reliance on human annotations, none of these approaches have leveraged the cross-attention coverage and focus signals as a direct proxy for evaluating the quality of generated responses. In traditional RLHF pipelines, external human preference data is used to train reward models which then guide the fine-tuning of the language model; however, these methods often entail significant human effort and cost. In contrast, our approach exploits the rich and naturally occurring information contained in the attention distributions of the model, enabling us to compute a reward function that captures essential qualities such as prompt coverage, attention sharpness, and repetition avoidance. This intrinsic alignment signal offers a novel perspective compared to preference-based or likelihood-based methods, providing a scalable and cost-effective route for fine-tuning large language models.

\section{Conclusion}

In this work, we introduced Cross-Attention--Guided Self-Reinforcement (CAGSR), a reinforcement learning framework designed to post-train large language models without relying on human feedback. Our method harnesses the cross-attention patterns generated by Transformer architectures as a self-supervised signal, which is used to formulate a reward function that measures the model's ability to maintain relevance, coherence, and appropriate focus on the input prompt. By embedding this reward function into a reinforcement learning loop, implemented using policy gradient methods such as Proximal Policy Optimization (PPO) \cite{schulman2017proximal}, we iteratively update the model's policy so that it progressively generates higher-quality outputs. 

Empirical results from our experiments show that CAGSR significantly improves performance in terms of prompt relevance and output quality compared to a baseline that uses no reinforcement learning as well as to a method that relies on synthetic preference signals. Although the performance still lags somewhat behind models fine-tuned using fully human-supervised RLHF, our approach represents an important step toward reducing the dependency on large-scale human annotation. Looking forward, future work should explore hybrid strategies that combine cross-attention-based self-reward with limited human feedback to further enhance the robustness of alignment. Additional avenues of research include extending our framework to more complex scenarios, such as chain-of-thought reasoning, code generation, and multi-modal tasks, and refining the reward function through the incorporation of interpretability or knowledge-grounding checks. By reducing the need for extensive human labeling while still maintaining a high level of model quality, CAGSR offers a scalable and cost-effective alternative for aligning large language models.

\bibliographystyle{plain} 
\bibliography{references} 
\end{document}